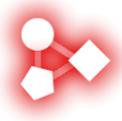

# Ethical Artificial Intelligence - An Open Question

**Alice Pavaloiu**
*Paris School of Business, France*
*(Graduate Student)*
*ada.staicu@gmail.com*

**Utku Kose**
*Usak University,*
*Computer Sciences App. & Res. Center, Turkey*
*utku.kose@usak.edu.tr*

## Abstract

Artificial Intelligence (AI) is an effective science which employs strong enough approaches, methods, and techniques to solve unsolvable real-world based problems. Because of its unstoppable rise towards the future, there are also some discussions about its ethics and safety. Shaping an AI-friendly environment for people and a people-friendly environment for AI can be a possible answer for finding a shared context of values for both humans and robots. In this context, objective of this paper is to address the ethical issues of AI and explore the moral dilemmas that arise from ethical algorithms, from pre-set or acquired values. In addition, the paper will also focus on the subject of AI safety. As general, the paper will briefly analyze the concerns and potential solutions to solving the ethical issues presented and increase readers' awareness on AI safety as another related research interest.

## 1. INTRODUCTION

The future depends on roles of many innovative fields in shaping the humankind's life. Artificial Intelligence (AI) is known as one of these fields as making the practical life more autonomous. This situation is because of rapid changes and developments in especially computer and electronics oriented technologies. In time, there have been many different innovations making revolutionary jumps in these technologies and except from some specific disadvantages, they have taken more and more role in the modern life, thanks to many different advantages – benefits. With active roles of such technologies, people now have a better life and looks hopeful to the future. But it is always an active question nowadays that what if the future would not be according to the humankind's desire. Such anxieties are generally associated with the existence of the humankind and other living creatures over the World. As impact of the AI can be measured greatly from both micro and macro-perspectives actively (Pavaloiu, 2016), it is also a remarkable research way to think about how can AI be used in an ethical manner and kept safely. Although it is something utopic in old times and can be imagined in science-fiction movies, AI is now the reality, which should be evaluated in a serious manner for a better future. Because of that, ethical AI has become a popular research interest in 21st Century.

The right-wrong dichotomy is no longer congruent with the rise of AI from an ethical angle. The pace and development of technological advancements bring concerns to tech experts regardless of its benefits in multiple fields. The European Union (2016) passed a draft report of a Motion for a European Parliament Resolution with recommendations to the Commission on Civil Law Rules on Robotics which outlines a possible legal framework with regards to the implications of AI unleashing the upcoming industrial revolution. The employment shift that may lead to economic disparity and wealth inequality, physical safety issues in case of system failure, misuse of technology, data protection, privacy issues and singularity concerns were acknowledged as reasons to create a regulatory control framework and establish a code of ethical conduct.

Other initiatives from the private sector include funding for robust and beneficial AI. Close relations are being developed with the academia and new grants are given to universities for the creation of ethical AI. Awareness is spread by panels held for debating moral dilemmas and ethical issues that may arise with AI advancements all over the world. Hence the focus falls onto AI safety, for solving a complex, ethical problem.

In the sense of the explanations, objective of this paper is to address the ethical issues of AI and explore the moral dilemmas that arise from ethical algorithms, from pre-set or acquired values. In addition, the paper will also focus on the subject of AI safety. As general, the paper will briefly analyze the concerns and potential solutions to solving the ethical issues presented and increase readers' awareness on AI safety as another related research interest. The authors believe that the paper is an effective reference work to anyone who is interested in the research focused on ethical and safety issues of AI.

Considering the main subject and research focus, remaining content is organized as follows: The next section is devoted to the concept of ethical AI by giving more emphasis on well-known concerns. In detail, this section takes readers interest to the heart of the ethical AI and enables them to have some idea about the associated literature. After that section, the third section explains some possible solutions for the mentioned ethical AI oriented concerns. Following the third section, the fourth section gives some brief information about AI safety as under the general scope of ethical concerns on AI. Finally, the paper ends with a discussion of conclusions and future work.

## 2. ETHICAL AI – CONCERNS

Nick Bostrom (2015) underpinned the importance of developing an AI which would not pose a threat to humanity, nor to its evolution. Accordingly, even if creating AGI already presents to be a challenge and creating people-friendly AGI would prove to be even more difficult, it is recommended to solve the second challenge first. Preventive measures need to be taken in order to insure AI safety where there could be unpredictable, unintended side effects of uncontrolled innovation. Notwithstanding the desire of an off-switch or undo button, the reversibility of a sequence of actions cannot be insured at the present moment. Consequently, Nadella (2017) recommended simulations in real, controlled environments in order to capture any unpredictable behavior and prevent any major crisis.

The future implications need to be considered not only by ethics committees or research departments, but also by the government, industry, international institutions and organizations at the earliest stages. The culture of responsibility needs to be developed at a global scale in order for an AI friendly environment to be created and maintained.

As can be expected, the progress of society at the technological level will not be stopped, but conducted by having instilled a code of ethics, with high precaution and a risk-management assessment anticipating all predictable outcomes. Nevertheless, the safety measures need to extend upon all the unpredictable actions that may occur from an autonomous form of thinking. Russell and Norvig (2010) pioneer the concept of robots needing to share the same values of people, whether they are instilled in their design or acquired, in order for safe AGI to be attained. A symbiotic relation between human beings and robots needs to be created, where both actors need each other to evolve and would rather work for a common goal, than turning against each other.

## 2.1. Moral Dilemmas Based On Pre-Set or Acquired Values

At the design level, a classic example is given by the three rules of Asimov (MIT Technology Review, 2016) for robots as tools or technology, which served as a foundation for engineering AI but did not answer the ethical issues.

1. *"A robot may not injure a human being or, through inaction, allow a human being to come to harm".*
2. *"A robot must obey orders given it by human beings except where such orders would conflict with the First Law".*
3. *"A robot must protect its own existence as long as such protection does not conflict with the First or Second Law. "*

The three laws created moral dilemmas of different types showing that even when a logical set of rules is applied, it is deemed to fail when interpreted by a different type of thinking. Therefore, giving autonomy to machines increased not only the benefits of technological advancements, but also the moral and legal implications.

A rigid set of values instilled by design led to an ethical predicament in the case of autonomous vehicles (Schoettle and Sivak, 2014). Should an autonomous vehicle for instance, when faced with an unavoidable collision with another car sacrifice the one passenger inside in order to save the other three passengers from the other car? 34.2% of people interviewed agreed to self-sacrifice but stated they would not buy an autonomous vehicle, despite the low probability of car crash and the increased rate of lives saved. The value of the individual life versus the greater good for the greater number generates a complex dilemma related to "Should a Driverless Car Decide Who Lives or Dies?" and the answer does not lie in the market (Naughton, 2015; Wallach, 2014).

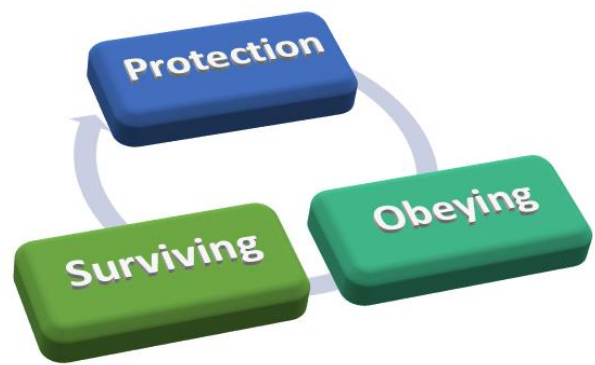

**Figure 1.** Three rules of Asimov for intelligent machines.

One solution would be to have an in-built ethical program that would align to the values of the car owner. Conversely, there is no consensus upon a set of universal values to follow, which may lead to another ethical issue.

Another type of dilemma that would arise with self-driving cars would be with the pre-set conditions to follow. Their moral capabilities and ethical sensitivity would need to be increasingly developed and refined in order for the vehicle to decide that breaking the speed limit at times might save the life of the passenger. The example of prioritizing under exceptional circumstances while minimizing the risk was given. Two noteworthy cases were analyzed from an ethical point of view. The one of a female passenger in labour and the second one of a severely wounded passenger, both lives or wellbeing depending on how fast the car could reach the hospital. Being able to recognize critical situations and prioritize the life of the passenger over speed limit could help save a life even at the expense of breaking the rules as a minimal damage.

Furthermore, ethical decision-making process would become more complicated where empathy and care are required. An elderly-care taker robot with a rigid adherence to the first law of Asimov would not be able to take a decision if the elder would not want to take his pills for instance. The robot would have to determine which choice, whether to allow the elder to skip the dose or to make him take the pills against his will, would not harm the elder. In this case, both courses of action would lead to harming the elder to a different extent.

Another approach is related to the top-down approach for solving this type of dilemma, by giving the robot more flexibility in reasoning. It would then be compelled to compute all the consequences of both courses of action and their consequences, in order to decide which is the lesser evil (Allen *et al.,* 2005; Wallach and Allen, 2008).

Another solution would be the bottom-up approach which enables the robot to learn from experience, instead of being given pre-set conditions to follow. Acting on habituated responses would allow the robot to develop moral behavior organically depending on the environment and its interaction with it.

Cassell (2016) proposed a different approach of human-robot interaction based on entrainment, the mirroring behavior that establishes between two actors who converse. If systems are taught how to entrain, they would come to act like the other conversational-agent they are interacting with. The findings of Breazeal (2002) in two distinct human-robot regulation and entrainment studies showed that both people and the anthropomorphic robot entrained by using expressive cues in the conversational process. The quality of interaction was improved for both, the human and the robot. One danger that could take place is social valence, which consists in affective mirroring from the human's perspective. The person would overestimate the ability of the robot to connect on a social and even emotional level. In order to maintain the bond, the person would become vulnerable to the robot "needs", acting unmindful and allowing potential for-profit actors who owned or manufactured the robot to take advantage of the situation.

Another method for infusing AI with an ethical conduct was brought by Russell and Norvig (2010), by inverse reinforcement learning, which consists in external awareness (Figure 2 and Figure 3 presents both reinforcement learning and inverse reinforcement learning algorithmic flows). The robot would observe another being's or another robot behavior and would try to determine what his/her objective is, without being computationally strained. Moral conduct could then be learnt by the robot through external input i.e. media, by understanding the basis of human emotion and the objective behind it, i.e. what is happiness, what makes people happy and why do people need to be happy. This action can lead to developing moral behavior or virtue ethics for the robot successfully.

## 2.2. Other Drawbacks of Pre-Set Values

Nevertheless, none of the approaches is unassailable. One drawback illustrated by Russell and Norvig (2010) is that by instilling values directly, engineers could omit or disregard an aspect which would later create a loop and lead to system failure. Most basic common-sense guidelines can often be overlooked by the engineers while designing the system which can lead to substantial errors in its modus operandi once deployed. AI can fail in ways people do not expect them at ordinary tasks, due to the different patterns of thinking.

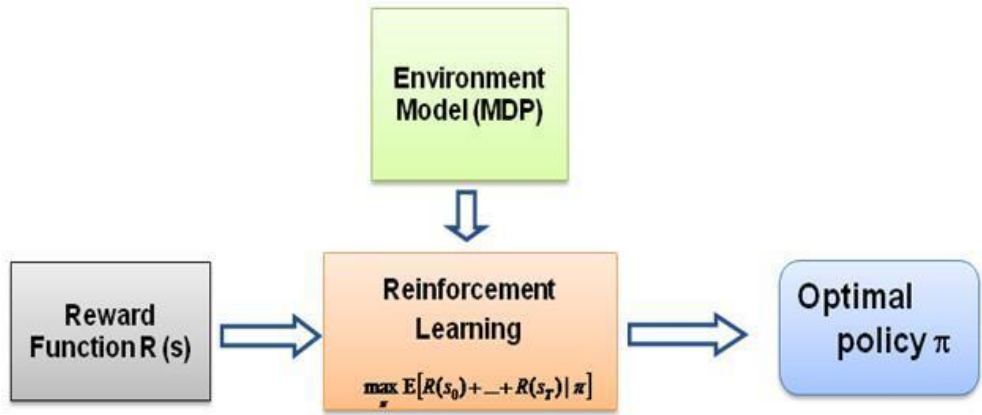

**Figure 2.** Reinforcement learning (Cornell University, 2011).

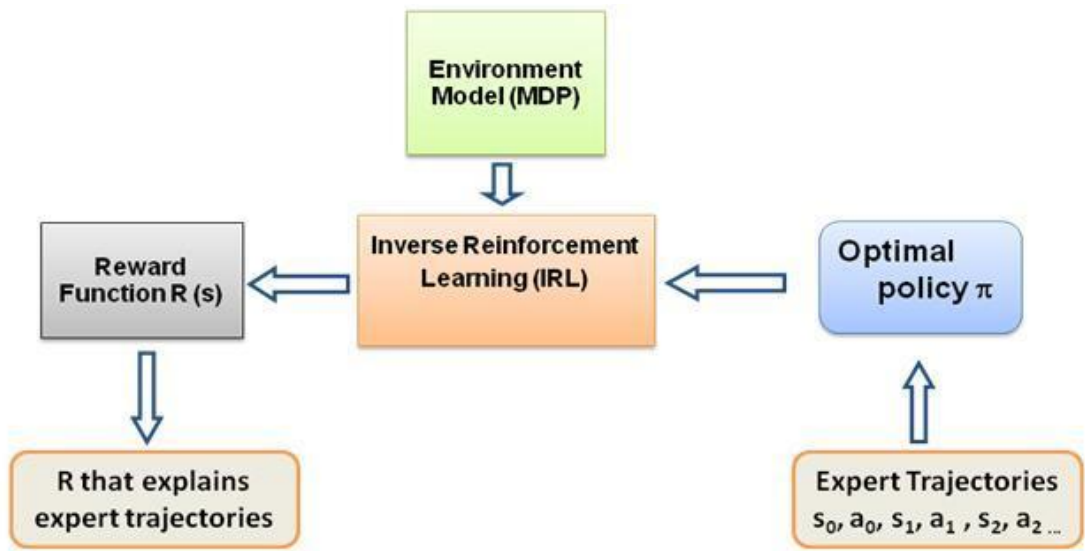

**Figure 3.** Inverse Reinforcement learning (Cornell University, 2011).

A colossal ethical issue was raised regarding functional errors in autonomous weapons. What is the impact of an error in regards to lethal fully autonomous weapons? Due to a human factor error at the level of programming, by either overlooking a basic rule or taking it as given, it could have irreversible effects and long-lasting consequences i.e. unintentionally start a war.

Nevertheless, good programming does not imply good judgement in warfare. Deeper moral issues arise in combat with regards to empathy or the lack of it. The military advantage has to be more significant than the potential civilian cost. Ethical issues with identifying the context for the military advantage or selecting a target in a group without collateral damage will change the nature of war. Another issue is represented by the autonomous weapons race which will lead to their proliferation and lower the barriers for more aggressive actors for entering a war, whether they are countries, de facto states, or terrorist groups (Lindh, 2016; Wallach and Allen, 2008).

The United Nations proposed to create an arms control agreement to ban autonomous weapons from being used without people being present at the decision-making process (Morris, 2016). From an ethical perspective, human judgement is different from machine judgement. The question is how much moral responsibility can people outsource to machines.

## 2.3. Other Drawbacks of Acquired Values

The benefits of acquired values are the higher-quality AI decisions which are congruent with the decisions taken by people. The similarity in judgement can lead to less errors and a better value alignment. Nevertheless, there are several drawbacks.

One ethical issue which sparked controversy is based on subjectiveness and hidden bias. Tufekci (2016) illustrated the afflicting situation where AI needs to take subjective decisions based on its increasing predictive power by answering open-ended questions. The example of a hiring algorithm was given. Due to the high levels of accuracy in prediction, an AI system can infer the likelihood of depression before symptoms appear only by analyzing the candidate's social media. The system could predict the likelihood of a potential candidate to get pregnant or choose a rather aggressive person who would fit in the corporate culture.

Hidden bias can also prove to be a substantial ethical issue when hiring. If an AI system learns through inverse reinforcement learning or any other method of acquiring ethical values organically, the hiring algorithm can reflect the corporate culture and amplify the already existing biases. Google algorithms show women less ads for high-paying jobs, while searching for an African-American name generates results with a criminal record (Tufekci, 2016). The dangerous nature of hidden biases can affect not only the potential candidates, but the company itself which may or may not be aware of the black-box algorithms. Thus transparency is one critical issue. The more powerful a predictive system is, the less transparent it is.

Notwithstanding, the high accuracy levels of prediction bring benefits in a similar manner. In medicine, preventive measures can be taken by a patient before the symptoms arise, which leads to an increasingly life-saving rate. According to different industry domains, if the prediction power is not ethically handled it can result in costs outweighing the benefits i.e. insurance companies can make use of the data for-profit in an unethical way, by over-charging people prone to different types of illness.

One solution would be to audit the black-boxes algorithm in order to discover and correct any hidden bias for a well-functioning system. Cognitive enterprises need to respect a regulatory framework and a code of ethics. One principle of the Code of Ethical Conduct for Robotics Engineers from the European Union charter proposed is inclusiveness (European Union, 2016). Transparency needs to insured by engineers. The right of access to information is inclusive to all stakeholders, who can also participate in the decision-making processes regarding the robotics research actions and activities, provided they are concerned by or involved in it.

## 2.4. Employment Shift - Economic Disparity - Wealth Distribution

An employment shift is generated with every new technological revolution. The job market changes. Labour is reduced by the jobs which are automated and a set of new, secondary jobs is created instead. A study published by Oxford University (2016) states that 47% out of the American existing jobs are to be replaced due to automation by 2043, while 69% out of the existing jobs in China and 75% of the ones in India will be substituted in the future. The European Union Commission (2016) forecasts that 90% of the jobs will have as a criteria basic digital skills. The new revolution will not only produce an employment shift, but also increase the wealth gap and lead to economic disparity.

One highly-debated solution is the universal basic income which would act as a buffer. A classic example was given by the following analogy. With Roomba arrival, the floor cleaning hover, the time spent for cleaning did not decrease, but remained constant. The only changing variable were the cleaning standards which increased. By comparison, once the basic income will be distributed, the time spent working will not decrease, but the standards of the job market will be upgraded.

The new paradigm that will allow people not to work for an income any more, but for pleasure instead, will lead to talent harnessing, new ways of contribution to society and finding meaning in activities which do not require labour.

The public and private sector would need to work together. New kinds of skills will emerge. Several skills that would insure success would not be all high-degree skills, i.e. teaching, nursing, creative jobs. IBM's CEO illustrated the new skill sector that will arise as the "new collar" (Balakrishnan and Lovelace Jr., 2017). Several initiatives of IBM develop the new pathway of technology by providing a relevant curriculum and making the fundamental changes in education in the data-driven economy.

The educational system needs to become more dynamic and flexible, especially time-wise. Technological advancements work at a computer pace, instead of human time. Implementing a relevant curriculum compatible with the new collar skills and jobs will help the transition in the employment shift greatly. Students need to be taught how to learn instead of how to respect authority and memorize information.

With the help of AI, people will be able to successfully do their job assisted where needed. The managers and executives will be able to focus on the top priorities and complex issues that cannot be solved by AI, instead of spending their time working on tedious tasks. The people whom jobs were automated will be trained for gaining a new set of skills compatible with the jobs of the future and would be able to maintain their lifestyle due to the basic income.

The new ecosystem of jobs will allow the culture of collaboration to flourish, preserve resources and maintain the balance of the systems put in place. If there will be inclusive growth, the GDP will raise with the help of AI and manage wealth disparity.

## 3. SOLUTIONS TO ETHICAL ISSUES

Morality varies across culture at a societal level. It is continuously morphing according to different trends, tendencies and technological advancements. Ethical issues become more and more complex with the new category AI comes with. A consensus is not likely to be reached from the right-wrong dichotomy. The solution resides in the complex nature of ethics. Machine objectives should align to people's objectives through values and ethical conduct (Russell and Norvig, 2010). Whether the values and morals are imparted at the programming level or acquired by learning and observing, no method is infallible. A more flexible work-frame at the design level could end up saving a life i.e. an autonomous vehicle being able to break the speed limit in case of an emergency while minimizing the damage. Another solution would be for people to take responsibility and be accountable for the irreversible effects AI can have whether it has functional, overlooked errors or it is misused i.e. fully autonomous weapons fallen on the wrong hands.

The system needs to be given an ethical code of conduct upon which to start basing its decision-making process. Prior to teaching AI how to be ethical, people need to think about what makes themselves ethical. Bad actors can become a real threat for the national security and may be able to release a course of action with long lasting consequences. The same outcome may be generated by corner cutting, not only by the misuse of technology. Regardless of time saving or cost reducing at the engineering or deployment level, corner cutting should be avoided. AI safety should be a top priority.

The new industrial revolution will not only focus on imparting morals onto machines, but bring a shift in human values in an equal manner. The EU charter illustrates in the Code of Conduct the human rights robots or any form of AI should respect and protect at all times – safety, privacy and dignity. AI should act for the benefit of humans and do not harm them in any way. Emergent behavior, which consists in the unpredictable rate of complex actions regardless of the ethical criteria a robot might base his decision upon, should strive for the minimal harm if unavoidable.

"The Human (IT) Manifesto" comprising of seven declarations for privacy, consent, identity, ability, ethics, good and democracy was circulated at the World Economic Forum (2017). The human being is central to the new industrial revolution for machines. The interaction between humans and AI is essential in order for a symbiotic relation to be created and for both parties to evolve with the help of the other. Entrainment is one way for improving the human robot interaction with the danger of social valence, the tendency of people to anthropomorphize AI and place too much trust in a human-like machine. The more people understand how interpersonal relationships work between the two different categories, the more people and robots can collaborate efficiently.

Morality cannot be outsourced to machines, even if there is algorithmic accountability for it. Erasing hidden bias is one goal towards a more transparent, trustworthy system and a healthy corporate culture. Audits and algorithm scrutiny are ways of preventing or correcting black-boxes algorithms (Nadella, 2017). Furthermore, real-time simulations in a controlled environment can bring benefits to the design of ethical AI. Developing human-friendly AGI algorithms without the power of applying decisions can lead to a deeper understanding of AI thinking, discovering new risks and potential benefits and preventing any irreversible crisis.

IBM principles for the AI revolution are based on skill building and retraining to insure an easier transition for the employment shift (World Economic Forum, 2015). Creating new skills compatible with the AI era will help create and maintain a healthy job system. Another principle is cross industry partnerships between engineers, academia, government and organizations. The partnerships would help engineers design better robots according to ethics, rules and regulations and potential societal or psychological factors that may affect human beings. Government and organizations would understand at the deeper level the functioning mechanisms of an AI which would allow the creation of a set of more accurate rules and regulations in conformity with the design (World Economic Forum, 2015; World Economic Forum, 2016).

## 4. AI SAFETY AS AN IMPORTANT FACE OF THE MADELLION

When considering the ethical AI, different rapidly growing research interests connected with the possibility of harmful effects of AI have also raised. These are generally because of anxieties on creating uncontrolled, not very well trained, or "very well trained" AIs, which are able to be cause of dangerous or harmful results affecting people's life and the existence of humankind or living organisms. Called as AI safety, this research interest is often structured over achieving the best AI systems that are friendly for people. Because AI is a creation of the humankind, it is always thought that such creation may employ humankind's mistakes, which can show "butterfly effect" at the end while solving important real world problems.

When the associated literature is examined, it can be seen that the concept of AI safety engineering was already proposed (Yampolskiy, 2013). Here, Yampolskiy (2013) argues that the understanding of "machine ethics" is having some mistakes – misguided ways and so, a new scientific approach of safety engineering is proposed for the related literature. Such ideas in the literature has led the researchers to think more about ensuring safe intelligent systems and more emphasis has been given to developing safe "agents" and safe approaches as shaping the edges of the AI safety.

According to the ethical AI, AI safety is associated with the technical, applied side of the same research road. According to the ethical AI, AI safety claims that the humankind should take care of all possible dangerous sides of AI in addition to its benefits and prevent possible negative effects by considering the "safety" in both philosophical and applied manner. Currently, the main focus of AI safety seems related with behaviors of AI systems as some kind of intelligent agents and have interest more in trained AI techniques. By eliminating details, some remarkable subjects that AI safety is currently dealing with can be mentioned briefly as follows:

- Interruptible Agents / Ignorant Agents / Inconsistent Agents / Bounded Agents (Evans and Goodman, 2015; Evans *et al.*, 2015; Orseau and Armstrong, 2016),
- Rationality (Lewis *et al.*, 2014; Russell, 2016),
- Corrigibility (Soares *et al.*, 2015),
- Reinforcement Learning / Inverse Reinforcement Learning (Abbeel and Ng, 2011; Ng and Russell, 2000),
- Adversarial Examples (Goodfellow *et al.*, 2017).

For both ethical AI and AI safety, "superintelligence" (Bostrom, 2014; Bostrom *et al.*, 2016) is also a remarkable common subject. Furthermore, the interest of future of AI and some connected interests like singularity (Goertzel, 2007; Kurzweil, 2005) and existential risk (Bostrom, 2002; Yudkowsky, 2008) are also other important factors shaping the whole literature of AI and its ethical – safe – future sides.

## 5. CONCLUSIONS AND FUTURE WORK

AI can solve the unsolvable problems. People need to be part of its development in order for AI to help augment human ingenuity and create a culture based on collaboration. There is no consensus about social norms, nor about the ethical principles AI should follow at the present moment. Until ethics becomes a vital part of human behavior, there is no very well answer to AI safety. But nowadays, there is also a remarkable interest in ensuring AI safety.

Shaping an AI-friendly environment for people and a people-friendly environment for AI can be a possible answer towards finding shared context of values for both humans and robots. The process of teaching machines to be more human-like may have a positive impact upon humans and translate into people becoming more human-like themselves. Henceforward, humanity will have the chance to realign its values accordingly, in addition to changing and enhancing their ethical conduct and to rethink their contribution to society at a deeper level.

### 5.1. Further Subjects to be Focused on

The issues of privacy, liability, rules and regulations, human rights and robot rights, superintelligence, and AI safety interests like specific (i.e. interruptible, ignorant) agents, rationality, corrigibility, and reverse learning were not addressed in detail under this research paper due to the broadness of the selected topic. On the other hand, the paper has not been formed on technical details for especially AI safety oriented problems. In this context, research made on "agents" is an important research way to be considered. Finally, some additional reading (extended bibliography) has been given after the section of References, in order to give interested readers to move forward more specifically about further theoretical and applied subjects.

## 5.2. Future Works

The authors are highly encouraged about moving more research works regarding to ethics of AI and AI safety. In this sense, some future applied works has been planned. These works include providing prototypes of some designed ethical models of intelligent agents and additional theories that can be introduced over applied approaches will be introduced to the associated literature in next reports.

**Additional Reading – Extended Bibliography**